\documentclass[10pt,twocolumn,letterpaper]{article}

\usepackage{cvpr}
\usepackage{times}
\usepackage{epsfig}
\usepackage{graphicx}
\usepackage{amsmath}
\usepackage{amssymb}
\usepackage{verbatim}
\usepackage{multirow}
\usepackage{array}

\usepackage[pagebackref=true,breaklinks=true,letterpaper=true,colorlinks,bookmarks=false]{hyperref}

\cvprfinalcopy 

\def\cvprPaperID{7} 

\ifcvprfinal\fi
\begin{document}

\title{IFQ-Net: Integrated Fixed-point Quantization Networks for Embedded Vision}

\author{Hongxing Gao, Wei Tao, Dongchao Wen\\
Canon Information Technology (Beijing) Co., LTD\\
{\tt\small \{gaohongxing,taowei,wendongchao\}@canon-ib.com.cn}
\and
Tse-Wei Chen, Kinya Osa, Masami Kato\\
Device Technology Development Headquarters, Canon Inc.\\
{\tt\small twchen@ieee.org}
}

\maketitle
\thispagestyle{empty}

\begin{abstract}
Deploying deep models on embedded devices has been a challenging problem since the great success of deep learning based networks. Fixed-point networks, which represent their data with low bits fixed-point and thus give remarkable savings on memory usage, are generally preferred. Even though current fixed-point networks employ relative low bits (\eg 8-bits), the memory saving is far from enough for the embedded devices. On the other hand, quantization deep networks, for example XNOR-Net and HWGQ-Net, quantize the data into 1 or 2 bits resulting in more significant memory savings but still contain lots of floating-point data. In this paper, we propose a fixed-point network for embedded vision tasks through converting the floating-point data in a quantization network into fixed-point.  Furthermore, to overcome the data loss caused by the conversion, we propose to compose floating-point data operations across multiple layers (\eg convolution, batch normalization and quantization layers) and convert them into fixed-point. We name the fixed-point network obtained through such integrated conversion as Integrated Fixed-point Quantization Networks (IFQ-Net). We demonstrate that our IFQ-Net gives 2.16$\times$  and 18$\times$  more savings on model size and runtime feature map memory respectively with similar accuracy on ImageNet. Furthermore, based on YOLOv2, we design IFQ-Tinier-YOLO face detector which is a fixed-point network with $256\times$ reduction in model size (246k Bytes) than Tiny-YOLO. We illustrate the promising performance of our face detector in terms of detection rate on Face Detection Data Set and Bencmark (FDDB) and qualitative results of detecting small faces of Wider Face dataset.

\end{abstract}

\section{Introduction}\label{sec:intro}



During the past decade, deep learning models have achieved great success on various machine learning tasks such as image classification, object detection, semantic segmentation, etc. However, applying them on embedded devices remains as a challenging problem due to the enormous resource requirement in terms of memory and computation power. On the other hand, fixed-point data inference yields promising reductions on such requirement for embedded devices~\cite{Fixed-pointEnergy}. Thus, fixed-point networks are primarily preferred when deploying deep models for the embedded devices.

In general, designing a fixed-point CNN network can be fulfilled by two types of approaches: 1)pre-train a floating-point deep network and then convert it into a fixed-point network; 2) train a deep CNN model whose data (\eg weights, feature maps, etc.) is natively fixed-point.
In~\cite{LinFixPoint}, a method is introduced to find the optimal bit-width for each layer to convert its floating-point weights and feature maps into their fixed-point counterparts. Given the hardware acceleration for 8-bit integer based computations,~\cite{NvidiaInt8} provides optimal thresholds which minimize the data loss during the 32-bits float to 8-bits integer conversion. These works have shown that it is feasible to significantly save memory usage through relatively low bit (\eg 8-bits) representation yet achieve similar performance. However, such memory saving is far from enough especially for embedded devices. The second approach is to train a network all of whose data is natively fixed-point. Nevertheless, as discussed in~\cite{LinOvercomming}, its training process may suffer from severe unstable weight updating because of the inaccurate gradients. Strategies such as stochastic rounding somehow result in improvement~\cite{FixTrain1,FixTrain2,FixTrain3} but a trade-off between low bit data representation and precise gradients still has to be made.


Alternatively, BinaryNet~\cite{BinaryConnect} employs binarized weights for forward pass but full precision weights and gradients for stable convergence. Meanwhile, its feature maps are also binarized to \{$-1$, $+1$\} so that its data can be represented as 1-bit fixed-point for less memory usage during inference time. 
However, a notable performance drop of 30\% (Top-1 accuracy) is observed on ImageNet classification. Subsequently, XNOR-Net~\cite{XNOR} employs extra scaling factors on both weights and feature maps so that their \lq\lq binary\rq\rq\, elements are generalized to \{$-\alpha$, $+\alpha$\} and \{$-\beta$, $+\beta$\} respectively. These extra factors enrich the data information thus gains 16\% accuracy back on ImageNet. Furthermore, HWGQ-Net~\cite{HWGQ} uses a more flexible k-bits quantization on feature maps whose elements can be generalized to \{0, $\beta$, $2\beta$, $3\beta$\} in the situation of 2-bits uniform quantization. Such $k$-bits feature maps ($k\geq2$) give a further 8\% improvement making HWGQ-Net to be the state-of-the-art quantization network on ImageNet classification.

Given a HWGQ-Net, each filter of its quantized convolution layer can be expressed as a multiplication of a floating-point $\alpha$ and a binary fixed-point matrix whose elements are limited to \{$-1$,$+1$\}. Similar representations can also be applied to its feature maps (see Equation~\ref{equa0}). Therefore, to obtain its fixed-point counterpart, it would be only necessary to convert the floating-point $\alpha$ and $\beta$ while other parts of the layer are natively fixed-point. Besides, Batch Normalization (BN) layer, which is usually employed on top of each convolution layer, also contains floating-point parameters and thus requires fixed-point conversion (see Equation~\ref{equ:BN0}). One way to do this is to separately convert each of the floating-point data but it usually results in data loss that would be accumulated over the network and cause a notable performance drop.

\begin{figure}[t]
\begin{center}
\includegraphics[width=0.9\linewidth]{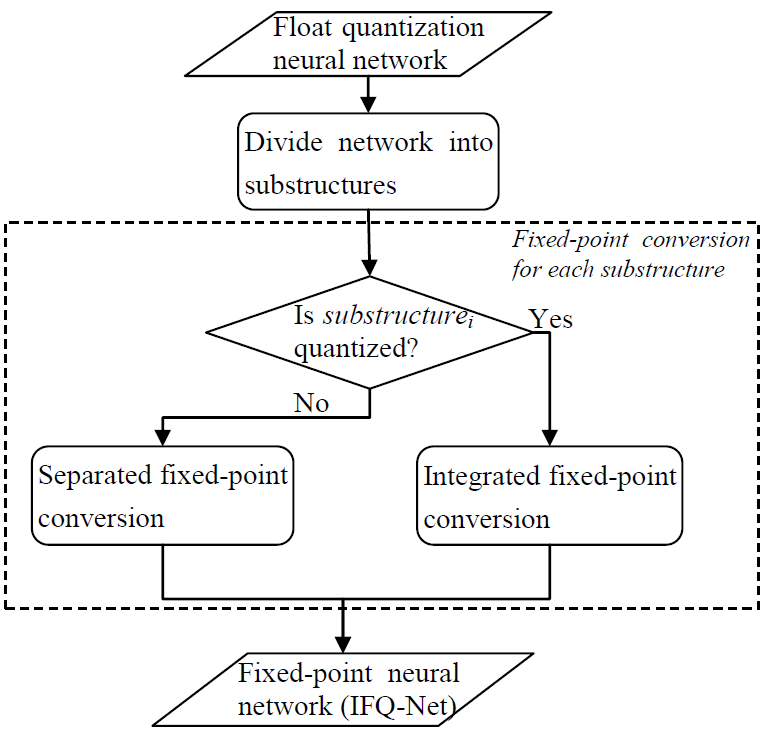}
\end{center}
\caption{The flowchart of converting a floating-point quantization network into IFQ-Net.}
\label{fig:flowchart}
\end{figure}

In this paper,  we propose a novel fixed-point network, IFQ-Net which is obtained through converting a floating-point quantization network into its fixed-point counterparts. As illustrated in Figure~\ref{fig:flowchart}, we first divide the quantization network into several substructures, where each substructure is defined as a group of consecutive layers that starts with a convolution layer and ends with a quantization layer. An example of the substructures of AlexNet is listed in Table~\ref{tab:AlexSubstructures}. Then we convert the floating-point data in each substructure into fixed-point data. Especially for the \lq\lq quantized substructure\rq\rq, which starts with a quantized convolution layer and ends with a quantization layer, we propose to compose its floating-point data into the thresholds of the quantization layer and then convert the composition result into fixed-point. As will be presented in section~\ref{intefixconv}, our integrated conversion method does not cause any performance drop. At the end, we separately convert each floating-point data in the remaining non-quantized substructures (if any) to fixed-point resulting in a fixed-point network, IFQ-Net. 

In this paper, the major contributions we made are:
\begin{itemize}
\item proposing IFQ-Net network, obtained through converting a floating-point quantization network into fixed-point. Due to the relatively low bits of the quantization network, IFQ-Net gives much more savings on model size and runtime feature map memory.
\item proposing an integrated conversion method to convert the floating-point data in the quantized substructures without performance drop. Since its BN operation (if available) is already integrated into the thresholds of the corresponding quantization layer, our IFQ-Net does not require actual BN implementation on target hardware. 
\item designing IFQ-Tinier-YOLO face detector, a fixed-point model with 256$\times$ tinier model size (246k Bytes) than Tiny-YOLO.
\item demonstrating the feasibility of quantizing all convolution layers in IFQ-Tinier-YOLO model, which differs from the original HWGQ-Net whose first and last layers are full precision.
\end{itemize}

\section{Quantized convolutional neural network}\label{Sec:QCNN}
A CNN network usually consists of a series of layers where the convolution layer monopolizes the inference time of the whole network. However, the weights and features maps were found redundant for most tasks. Consequently, enormous efforts have been done on quantizing the weights and/or the input feature maps into low-bit data for less memory usage and fast computation. 


\subsection{Quantization network inference}\label{subSec:QCNNInfere}
Embedded devices are usually employed for network inference only because of their limited computation resources. Hence, in this paper, we mainly focus on the inference process of a network. 
In the following, we take a typical quantized substructure from HWGQ-Net as an example to illustrate the computation details of its inference.


For the $l$-th convolution layer of HWGQ-Net, we use $\textbf{W}\in\mathbb{R}^{c\times h{'} \times w {'}}$ and $\textbf{X}^l\in\mathbb{R}^{c\times h \times w}$ to represent one of the filters and its input feature maps respectively, where $c$, $h{'}$, $w{'}$, $h$, $w$ are the number of channels, the height and width of its filter, and the height and width of the input feature maps respectively. 
In the case of a 2-bit quantized convolution layer from HWGQ-Net, its filter is binarized into $\textbf{W}\in\{-\alpha, +\alpha\}^{c\times h{'}\times w{'}}$ and $\textbf{X}^l\in\{0, \beta, 2\beta, 3\beta\}^{c\times h \times w}$. 
Then the computation of a convolution layer can be represented as

\begin{equation} \label{equa0}
\textbf{Y}^{conv}=\textbf{W}\otimes \textbf{X}^l+b=\alpha\beta \cdot\textbf{W}_b\otimes \textbf{X}^l_q+b
\end{equation}

where $\otimes$ represents the convolution operation; $\textbf{W}_b$ and $\textbf{X}^l_q$ are integer part of the quantized filter and feature maps so that $\textbf{W}_b \in \{-1,+1\}^{c\times h{'}\times w{'}}$ and $\textbf{X}^l_q \in\{0, 1, 2, 3\}^{c\times h \times w}$, $b$ is its learned bias.$ \textbf{Y}^{conv}$ is its output feature map. 


Typically, a BN layer is applied on top of a convolution layer. It is computed in an element-wise manner as follows,
\begin{equation} \label{equ:BN0}
\textbf{Y}^{BN}_{i,j}=\frac{\textbf{Y}^{conv}_{i,j}-\theta}{\sigma}
\end{equation}
where $0\leq i<w$, $0\leq j<h$, $\theta$ and $\sigma$ are the learned mean and variance of the feature map.

At the end, a quantization layer maps its input feature map $\textbf{Y}^{BN}$ into discrete numbers. Taking a 2-bits uniform quantization for instance, its computation can be expressed as
\begin{equation}\label{equ:quant0}
\textbf{X}^{l+1}_{i,j}=\beta{'}* \left\{
\begin{array}{cc}
0         &{      \textbf{Y}^{BN}_{i,j}  \leq  thr_1} \\
1      &{thr_1  <  \textbf{Y}^{BN}_{i,j}  \leq  thr_2} \\
2  &{thr_2< \textbf{Y}^{BN}_{i,j}  \leq  thr_3} \\
3   &{ \textbf{Y}^{BN}_{i,j} >thr_3}
\end{array}
\right.
\end{equation}
where $ thr_1$, $ thr_2$ and $ thr_3$ are the thresholds used for quantizing its input $\textbf{Y}^{BN}$, and $\beta{'}$ is the scale factor for its output feature map. The resulting $\textbf{X}^{l+1}$ is then employed as the input of the ($l+1$)-th convolution layer (if available).

When max pooling layer appears in the substructure, as discussed in \cite{XNOR}, it is better to place it between convolution and BN layers for richer data information. In other words,
\begin{equation}
\textbf{Y}^{pooling}_{i,j}=\max \limits_{(m,n)\in A_{i,j}}\{\textbf{Y}^{conv}_{m,n}\}
\end{equation}
where $A_{i,j}$ denotes the local zone employed for pooling operation at location ($i$, $j$) of $\textbf{Y}^{conv}$. Then the input of the BN layer in Equation~\ref{equ:BN0} is accordingly changed to be $\textbf{Y}^{pooling}_{i,j}$ .

\subsection{Separated fixed-point conversion} \label{sec:sepConv}
As illustrated in subsection~\ref{subSec:QCNNInfere}, the dominating part of the convolution computation $\textbf{W}\otimes \textbf{X}^l$ can be implemented with native fixed-point data only. However, the network still contains lots of floating-point data these being the scaling factor $\alpha$ and $\beta$ in the convolution layer, $\theta$ and $\sigma$ in the BN layer and also $thr_i$ in the quantization layer. Consequently, it is necessary to convert them into fixed-point when designing fixed-point networks for embedded devices.

\begin{figure}[t]
\begin{center}
\includegraphics[width=0.9\linewidth]{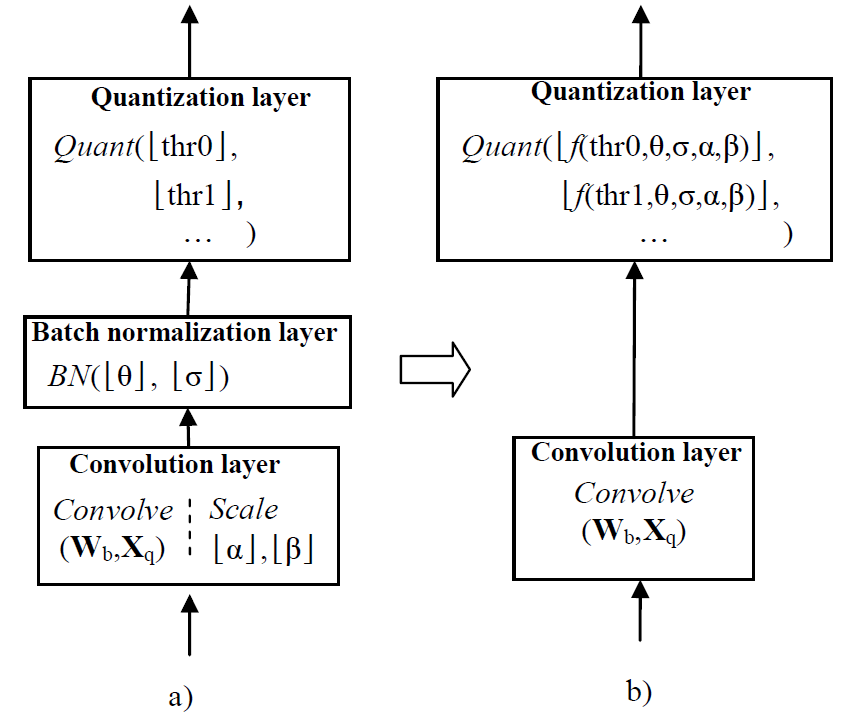}
\end{center}
\caption{Fixed-point conversion for a substructure: a) separated conversion which separately transforms each floating-point data into fixed-point data through the floor function $\lfloor\cdot\rfloor$ ; b) integrated conversion which employs a composition operation $f$ to compose all the floating-point calculations to quantization layer and then apply the fixed-point conversion for the composed results.}
\label{fig:fixpoint}
\end{figure}

A traditional way for the aforementioned conversion is to process them separately. As shown in Figure~\ref{fig:fixpoint}a, each floating-point data of the substructure is converted into its fixed-point counterpart. Since directly applying a simple conversion causes significant data loss especially when $\alpha$ is small (\eg 0.001), we use a relatively large $Q_m$ to scale-up the floating-point data\footnote{For fast calculation, $Q_m$ is usually set to $2^m$ so that the multiplication can be implemented by simple $m$-bit left shift}. For example, $\alpha$ can be transformed by $\lfloor\alpha Q_m\rfloor$ where $\lfloor\cdot\rfloor$ denotes the flooring operation. At the end, $Q_m$ has to be divided back to achieve \lq\lq equivalent\rq\rq\, outputs. Then, fixed-point conversion of a quantized convolution layer can be expressed as 


\begin{equation}
\textbf{Y}^{conv}=\frac{\lfloor\alpha Q_{m}\rfloor \lfloor\beta Q_{m}\rfloor\cdot\textbf{W}_b\otimes \textbf{X}_q+\lfloor b Q^2_{m}\rfloor}{Q^2_{m}}
\end{equation}

To obtain a substructure with fixed-point data only, the same conversion $\lfloor\cdot\rfloor$  is also applied to $\theta$, $\sigma$, $thr_i$ separately.

\section{IFQ-Net methodology}

To obtain a fixed-point network for embedded devices, we propose to first train a quantization network and then convert its floating-point data, which has been quantized into extremely low bits (\eg 1 or 2 bits), into fixed-point data. As demonstrated in Figure~\ref{fig:flowchart}, our methodology consists of two steps: first we divide a trained floating-point quantization network into substructures and then we convert each substructure into its fixed-point counterpart. We employ HWGQ-Net algorithm to train a floating-point quantization network.


\subsection{Substructure division}

\begin{figure}[t]
\begin{center}
\includegraphics[width=0.8\linewidth]{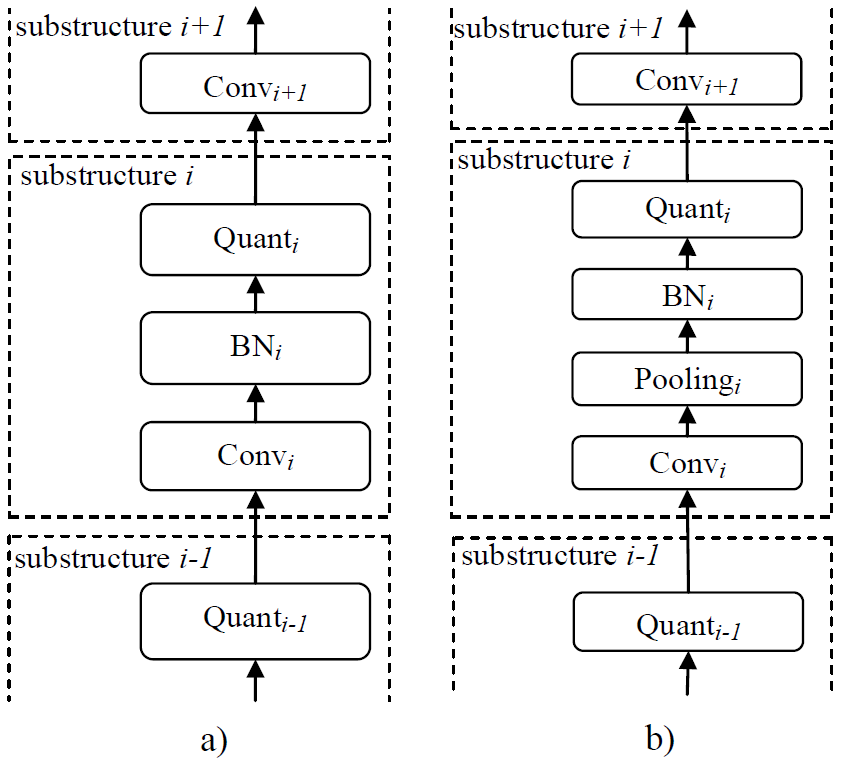}
\end{center}
\caption{Substructure division for a quantized network: a)substructure without max pooling layer; b)substructure with max pooling layer.}
\label{fig:substructure}
\end{figure}

As mentioned in Section~\ref{sec:intro}, a substructure is defined as a group of consecutive layers that starts with a convolution layer and ends with a quantization layer. Given a quantization network, we search for the quantized substructures in the network as demonstrated in Figure~\ref{fig:substructure}. Typically, the architecture of a quantized substructure is either \{convolution, BN, quantization\} or \{convolution, pooling, BN, quantization\}. The substructures that contain more than one convolution or quantization layer are not considered as quantized substructures. The layers between quantized substructures are defined as non-quantized substructures, which will be treated differently during fixed-point conversion. Generally, BN and/or max pooling layers are placed between convolution layers and quantization layers.  


Taking AlexNet-HWGQ network as an example, we divide it into 7 substructures (see Table~\ref{tab:AlexSubstructures}). Because the HWGQ network keeps its first and last convolution layer full precision, so the corresponding substructures ($substructure1$ and $substructure7$) are non-quantized and thus will be converted differently. Please note that we group all the layers on top of the $FC7$ layer as one single substructure.


\subsection{Integrated fixed-point conversion} \label{intefixconv}

A trained quantization network can be divided into substructures that contain lots of floating-point data. To obtain a fixed-point network, it is necessary to convert each of its floating-point substructures into fixed-point. However, converting the floating-point data in a separated manner usually leads to performance drop. Consequently, in the following, we introduce an integrated way to convert a floating-point substructure. Taking 2-bits uniformly quantized substructure from HWGQ-Net as an example, its computations that mentioned in Equation~\ref{equa0},~\ref{equ:BN0} and~\ref{equ:quant0} can be composed as follows

\begin{equation}\label{equ:integrated1}
\textbf{Y}^{quant}=  \left\{
\begin{array}{rl}
0         &{     \frac{\alpha\beta \cdot\textbf{W}_b\otimes \textbf{X}_q+b -\theta}{\sigma}  \leq  thr_1} \\
\beta{'}      &{thr_1  <  \frac{\alpha\beta \cdot\textbf{W}_b\otimes \textbf{X}_q+b -\theta}{\sigma}  \leq  thr_2} \\
2\beta{'}     &{thr_2< \frac{\alpha\beta \cdot\textbf{W}_b\otimes \textbf{X}_q+b -\theta}{\sigma}  \leq  thr_3} \\
3\beta{'}     &{ \frac{\alpha\beta \cdot\textbf{W}_b\otimes \textbf{X}_q+b -\theta}{\sigma} > thr_3}
\end{array}
\right.
\end{equation}
Since $\alpha>0$, $\beta>0$ and also $\sigma>0$, Equation~\ref{equ:integrated1} can be transformed to
\begin{equation}\label{equ:integrated2}
\textbf{Y}^{quant}=  \beta{'}*\left\{
\begin{array}{rl}
0         &{                                          \textbf{W}_b\otimes \textbf{X}_q \leq   \frac{thr_1*\sigma+\theta-b}{\alpha\beta}} \\
1      &{\frac{thr_1*\sigma+\theta-b}{\alpha\beta}    <  \textbf{W}_b\otimes \textbf{X}_q  \leq  \frac{\sigma*thr_2+\theta-b}{\alpha\beta}} \\
2     &{\frac{\sigma*thr_2+\theta-b}{\alpha\beta} <  \textbf{W}_b\otimes \textbf{X}_q  \leq  \frac{\sigma*thr_3+\theta-b}{\alpha\beta}} \\
3     &{                                          \textbf{W}_b\otimes \textbf{X}_q > \frac{\sigma*thr_3+\theta-b}{\alpha\beta}}
\end{array}
\right.
\end{equation}

As illustrated in Equation~\ref{equ:integrated2}, all the floating-point data of a quantized substructure is composed into the newly formed thresholds (\eg $\frac{thr_1*\sigma+\theta-b}{\alpha\beta}$). Such composition process is performed with floating-point data and thus does not impact the output result.

The next step is to convert the new thresholds into fixed-point data. $\textbf{W}_b$ and $\textbf{X}_q$ are both integers thus the resulted $\textbf{W}_b\otimes \textbf{X}_q$ are also integers. In Equation~\ref{equ:integrated2}, when thresholding the integers $\textbf{W}_b\otimes \textbf{X}_q$ with newly formed floating-point thresholds, theoretically, their fractional parts do not affect the result. Consequently, we can simply discard the fractional part by applying the floor function $\lfloor\cdot\rfloor$ on the new thresholds. Compared to the separated fixed-point conversion method, our method does not require to scale-up the floating-point data with $Q_m$ yet gives identical quantization results. Besides, the remaining floating-point data $\beta{'}$ can be processed in the next substructure just like how we deal with the $\beta$ of Equation~\ref{equa0}. Consequently, all the computations of a quantized substructure, as represented in Equation~\ref{equ:integrated2}, can be casted on fixed-point data after $\lfloor\cdot\rfloor$ is applied on each of the new thresholds.

The fixed-point implementation of on-device BN computation is challenging for embedded devices. As an alternative solution which differs from the method that merges it into a convolution layer, the proposed integrated fixed-point conversion method transforms the BN computation into the new quantization thresholds. Consequently, our IFQ-Net also does not require actual BN implementation on embedded hardware.


In the above, we have taken $k=2$ as an example of converting a $k$-bits quantization network. However, when a larger $k$ is employed, it would be necessary to store $(2^k-1)$ thresholds. In the uniform quantization scenario, the network's thresholds can be expressed as $thr_i=i*base + offset$. Thus, one may only need to store $base$ and $offset$ because all the thresholds can be restored from them. Similarly, denoting $base{'}=\frac{\sigma*base}{\alpha\beta}$ and $offset{'}=\frac{\theta - b + \sigma*offset}{\alpha\beta}$, our newly formed thresholds can also be represented as $thr{'}_i=i*base{'} + offset{'}$. Thus, our new thresholds $thr{'}$ can also be represented in an efficient way.  
Then computations in a k-bits uniformly quantized substructure can be expressed as Equation~\ref{equ:integrated3} which can be further converted into fixed-point in an integrated manner.
\begin{equation}\label{equ:integrated3}
\textbf{Y}^{quant}= \beta{'}*\left\{
\begin{array}{cc}
0         &{                                          \textbf{W}_b\otimes \textbf{X}_q \leq  thr{'}_{1}} \\
1      &{thr{'}_{1}            <  \textbf{W}_b\otimes \textbf{X}_q  \leq  thr{'}_{2}} \\
2     &{thr{'}_{2} <  \textbf{W}_b\otimes \textbf{X}_q  \leq  thr{'}_{3}} \\
\vdots     &   \vdots  \\
(2^k-1)     &{                                          \textbf{W}_b\otimes \textbf{X}_q > thr{'}_{(2^k-1)}}
\end{array}
\right.
\end{equation}

In summary, we presented IFQ-Net obtained by dividing a quantization network (\eg HWGQ-Net) into floating-point substructures and then converting each of them into fixed point. For the quantized substructures, we propose an integrated fixed-point conversion method which gives no performance drop. At the end, for the remaining non-quantized substructure (if any), we employ the separated method to convert them into fixed-point.


It is worth to point out that our IFQ-Net differs from the floating-point data composition method presented in~\cite{LayerMerge} in many aspects: 1)the paper claims that it combines multiple layers but does not explicitly explain how; 2)the paper applies the floating-point data composition for enabling binary convolution computation leaving other parameters as floating-point while our method is proposed for fixed-point conversion and 3)the paper concentrates on implementing a quantized network on FPGA but the performance (\eg detection rate, mAP etc.) is not reported.

\section{Experimental results}

In this section, we demonstrate how we convert each substructure of AlexNet into fixed-point to obtain an IFQ-AlexNet. We first test the performance of the integrated conversion method for the quantized substructures. Then, for the non-quantized substructures, we demonstrate how we experimentally set the scale factor $Q_m$ for the separated fixed-point conversion. We compare the performance of our IFQ-AlexNet with \lq\lq Lin~\etal~\cite{LinFixPoint}\rq\rq\, which is the state-of-the-art AlexNet-based fixed-point network on ImageNet. Furthermore, we also illustrate the performance of IFQ-Tinier-YOLO face detector which is an extremely compact fixed-point network on both FDDB and Wider Face datasets.


\subsection{IFQ-AlexNet network}

To obtain fixed-point networks, we first train floating-point quantization networks AlexNet-HWGQ whose weights and feature maps are quantized into 1-bit and $k$-bits ($k\in\{2,3,4\}$) respectively. The AlexNet-HWGQ is trained with 320k iterations on ImageNet while the batch size is set to 256. The initial learning rate is set to 0.1 and decreased by a factor of 0.1 every 35k iterations. We inherit other training settings from~\cite{HWGQ} and achieve similar performance.


\begin{table}[!ht]
\centering
\caption{Substructures of AlexNet-HWGQ network.} \label{tab:AlexSubstructures}
\setlength{\tabcolsep}{2.0pt}
\begin{tabular}{p{0.28\columnwidth}<{\centering}|p{0.28\columnwidth}<{\centering}|p{0.05\columnwidth}<{\centering}|p{0.28\columnwidth}<{\centering}}
  \hline
         {$substructure1$} &{$substructure2$} & {...} &{$substructure7$} \\
  \hline\hline
         Conv$_1$          & Conv$^q_2$      &\multirow{4}{*}{...} &{FC$^q_7$} \\
         Pool$_1$          & Pool$_2$         &                     &{BN$_7$} \\
         BN$_1$            & BN$_2$           &                     &{ReLU$_7$}\\
         Quant$_1$         & Quant$_2$        &                     &{FC$_8$} \\
  \hline
\end{tabular}
\end{table}

As the first step for obtaining the IFQ-AlexNet, we divide a floating-point AlexNet-HWGQ network into 7 substructures ( Table~\ref{tab:AlexSubstructures}). In the table, the superscript $q$ in Conv$^q_2$  and FC$^q_7$ means that their weights are binarized (1-bit) and input feature maps are quantized into $k$ bits by their bottom Quant$_i$ layers. We group the layers \{FC$^q_7$, BN$_7$, ReLU$_7$, FC$_8$\} as a single non-quantized substructure.

In the following, we will show how to convert each substructure into fixed-point to obtain an IFQ-AlexNet. In subsection~\ref{sec:4.1.1}, we show the performance of the proposed integrated conversion method for the quantized substructure while the non-quantized substructures are kept floating-point. We then illustrate the way to set a proper scaling factor $Q_m$ for converting each of the remaining non-quantized substructures (see subsection~\ref{sec:4.1.2}).  At the end, in subsection~\ref{sec:4.1.3}, we compare our IFQ-AlexNet with \lq\lq Lin~\etal~\cite{LinFixPoint}\rq\rq\, which is the state-of-the-art AlexNet-based fixed-point network.


\subsubsection{Integrated conversion for the quantized substructures} \label{sec:4.1.1}

In this subsection, we focus on converting the quantized substructures ($substructure2$,...,$substructure6$). The ImageNet Top-1 classification accuracy is employed to evaluate the accuracy of the converted networks (see Table~\ref{tab:fixdiffbits}). In the table, 
\lq\lq separated\rq\rq\, refers to the networks obtained by converting the quantized substructures of the corresponding AlexNet-HWGQ ($k$ equals to 2 or 3 or 4) in a separated manner (see section~\ref{sec:sepConv}). In contrast, \lq\lq integrated\rq\rq\, represents the networks obtained by converting their quantized substructures in the proposed integrated way (see section~\ref{intefixconv}). Please note that, to compare the performance of different conversion methods on quantized substructures, we keep the non-quantized substructures ($substructure1$ and $substructure7$) floating-point.


\begin{table}[!h]
\centering
\caption{Performance of different methods on converting the quantized substructures of AlexNet-HWGQ networks on ImageNet top-1 classification accuracy.}
\label{tab:fixdiffbits}
\begin{tabular}{c|c|c|c}

  \hline                               &$k=2$  &$k=3$  &$k=4$ \\
  \hline\hline     {AlexNet-HWGQ}             &0.5214 &0.5301 &0.5471 \\
   \hline          {separated($m=12$)}  &0.5206 &0.5296 &0.5470 \\
                   {separated($m=10$)}  &0.5168 &0.5292 &0.5443 \\
                   {separated($m=9$)}   &0.5073 &0.5230 &0.5385 \\
                   {separated($m=8$)}   &0.4585 &0.4678 &0.5105 \\
   \hline          {integrated($m=0$)} &\textbf{0.5214} &\textbf{0.5301} &\textbf{0.5471} \\
  \hline
\end{tabular}
\end{table}


As shown in Table~\ref{tab:fixdiffbits}, the floating-point AlexNet-HWGQ networks achieves competitive classification accuracy. However, \lq\lq separated\rq\rq\, method shows notable performance degradation. The reason is that it separately converts each floating-point data $x$ of a quantized substructures by $\lfloor x*Q_m \rfloor$ which leads to data loss. To reduce such loss, a large $m$ has to be applied ($m=12$) which in turn causes more memory usage. In contrast, for each quantized substructure, our \lq\lq integrated\rq\rq\, method gives identical outputs as its floating-point counterpart in AlexNet-HWGQ while the scaling factor $Q_m$ is not required at all ($m=0$). Even though we employ the uniform quantization as example, our \lq\lq integrated\rq\rq\, method is also effective for the networks quantized by other strategies as long as their floating-point operations can be composed as in Equation~\ref{equ:integrated2}.

\subsubsection{Separated conversion for the non-quantized substructures} \label{sec:4.1.2}

In the subsection~\ref{sec:4.1.1}, we have demonstrated that the proposed integrated method gives lossless fixed-point conversion for quantized substructures. To obtain IFQ-AlexNet all of whose data operations are fixed-point data based, we then convert each of the remaining non-quantized substructures in a \lq\lq separated\rq\rq\, manner. For saving more memory while causing less conversion loss for such substructures, an optimal $Q_m$ is required for each of non-quantized substructure: $substructure1$ and $substructure7$. Since the $substructure7$ directly outputs the inferred results for the task, the preciseness of its computation is more critical. Consequently, we first find the optimal $m$ for its fixed-point conversion while $substructure1$ is kept floating-point.

\begin{figure}[!ht]
\begin{center}
\begin{tabular} {@{}cc@{}}
\includegraphics[width=0.45\linewidth]{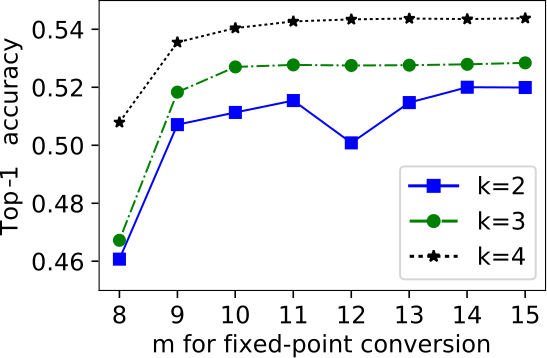} &
\includegraphics[width=0.45\linewidth]{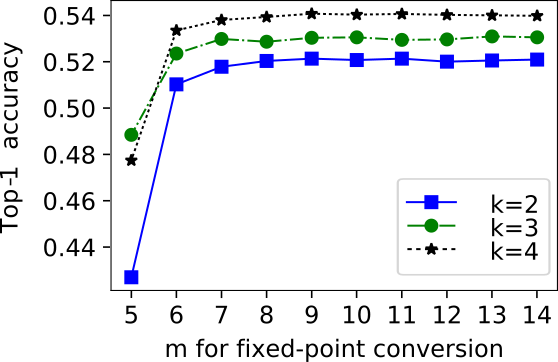} \\
 a):$subtructure7$ & b):$substructure1$
\end{tabular}
\caption{Top-1 accuracy on ImageNet of networks with various $m$ for $subtructure7$ and $substructure1$ fixed-point conversion.}
\label{fig:sub7}
\end{center}
\end{figure}

As demonstrated in Figure~\ref{fig:sub7}a), for the networks with different $k$, a larger $m$ for $Q_m$ generally gives better performance. It is because a larger $m$ value gives less data loss during each fixed-point conversion $\lfloor x*2^m \rfloor$. Nevertheless, when $m\geq14$, no further performance improvement can be observed for all the three networks indicating $m=14$ would be sufficient for fixed-point conversion for the floating-point data in $substructure7$.

By fixing $m=14$ for converting $substructure7$ into fixed-point, we then optimize the $m$ for $substructure1$. As shown in Figure~\ref{fig:sub7}b), $m=9$ can be considered as the sufficient scaling factor for the fixed-point conversion of $substructure1$.

 In summary, to obtain IFQ-AlexNet, we employ the lossless \lq\lq integrated\rq\rq\, conversion method for the quantized substructures and $m=9$ and $m=14$ for the scaling factor $Q_m$ for converting the  $substructure1$ and $substructure7$ of AlexNet-HWGQ networks respectively.


\subsubsection{Performance comparison} \label{sec:4.1.3}

In the following, we compare our IFQ-AlexNet with \lq\lq Lin~\etal~\cite{LinFixPoint}\rq\rq\, which is the state-of-the-art AlexNet-based fixed-point network. Lin~\etal~\cite{LinFixPoint} employ a $\gamma$ ($\gamma\geq 9$) as the number of bits for representing each data of the first layer and then introduce an optimal setting on the number of bits for other layers with respect to  $\gamma$ (see Table~\ref{tab:networkcomp}). It is worth to point out that \lq\lq Lin~\etal~\cite{LinFixPoint}\rq\rq\, is converted from an AlexNet-like network which posses $\sim$2$\times$ savings on the number of parameters compared to our IFQ-AlexNet (21.5 million vs. 58.3 million\footnote{To be consistent with the reference paper~\cite{LinFixPoint}, the parameters in $FC_8$ are not included.}).

Table~\ref{tab:networkcomp} compares the number of bits that are employed to represent every fixed-point data of each layer of \lq\lq Lin~\etal~\cite{LinFixPoint}\rq\rq\, and our IFQ-AlexNet. As shown in the table, for $conv2$$\sim$$conv5$ layers, IFQ-AlexNet employs 1-bit for representing their weights which is remarkably lower than \lq\lq Lin~\etal~\cite{LinFixPoint}\rq\rq\,. Most importantly, for $FC6$ and $FC7$ layers which are parameter intensive and thus dominate the model size,  we consistently employ 1-bit weights. Thus, our IFQ-Net gives 6$\times$ savings (1-bit vs. 6-bits). On the other hand, regarding to the feature maps, our IFQ-AlexNet networks also generally use lower bits than their competitors (the same bits may happen on $conv2$ and $conv4$ layers only if $k=4$ and $\gamma=9$).
\begin{table} [!h]
\centering
\caption{Comparison on the number of bits employed for each layer of AlexNet-based fixed-point networks.}\label{tab:networkcomp}
\setlength{\tabcolsep}{2pt}
\begin{tabular}{p{1.0cm}<{\centering}|p{1.2cm}<{\centering}|p{2.0cm}<{\centering}|p{1.2cm}<{\centering}|p{2.0cm}<{\centering}}

  \hline                   & \multicolumn{2}{c|}{Lin~\etal~\cite{LinFixPoint} ($\gamma\geq 9$)} &\multicolumn{2}{c}{IFQ-AlexNet}\\
   \cline{2-5}             &Weights & Feature maps &Weights & Feature maps \\
  \hline\hline   $conv1$   &$\gamma$     &$\gamma$   & 9     &8  \\
                 $conv2$   &$\gamma$-5   &$\gamma$-5 & 1     &k  \\
                 $conv3$   &$\gamma$-4   &$\gamma$-4 & 1     &k  \\
                 $conv4$   &$\gamma$-5   &$\gamma$-5 & 1     &k  \\
                 $conv5$   &$\gamma$-4   &$\gamma$-4 & 1     &k  \\
                 $FC6$     &6            &6          & 1     &k  \\
                 $FC7$     &6            &6          & 1     &k  \\
  \hline
\end{tabular}
\end{table}

For \lq\lq Lin~\etal~\cite{LinFixPoint}\rq\rq\, networks, different $\gamma$ give different preciseness of its fixed-point data. We directly borrow the experimental results from the paper setting $\gamma$ to 9 and 10. Table~\ref{tab:perfcompare} illustrates the memory usage of the weights and feature maps of the fixed-point networks in terms of millions of bits (Mbits). As shown in the table,  regarding to the model size of the compared fixed-point networks, our IFQ-AlexNet networks ($k$ = 2 or 4) give 2.16$\times$ savings(58.8 Mbits vs. 127.3 Mbits) over \lq\lq Lin~\etal~\cite{LinFixPoint} ($\gamma$ = 9)\rq\rq.

\begin{table} [!h]
\centering
\caption{Model size (Mbits), inference memory for feature maps (Mbits)  and performance comparison of fixed-point networks.}\label{tab:perfcompare}
		\setlength{\tabcolsep}{2pt}
\begin{tabular}{p{2.6cm}<{\centering}|p{1.0cm}<{\centering}|p{1.2cm}<{\centering}|p{1.0cm}<{\centering}|p{1.0cm}<{\centering}}
		\hline                   & \multicolumn{2}{c|}{Lin~\etal~\cite{LinFixPoint}} &\multicolumn{2}{c}{IFQ-AlexNet}\\
		\cline{2-5}              & $\gamma$ = 9  & $\gamma$ = 10 &$k$ = 2  & $k$ = 4 \\
		\hline\hline  {Model size}   &127.3     &128.5     & \textbf{58.8}     &\textbf{58.8}  \\
		\hline {Inference memory (feature maps) }  &10.8   &12.0 & \textbf{0.6}     &1.1  \\
		\hline {Top-5 accuracy}   &0.74  &\textbf{0.78} & 0.76     &\textbf{0.78} \\
		\hline
	\end{tabular}
\end{table}

To evaluate the memory usage of feature maps during inference time, we measure the maximum memory that consumed by one single layer, which is $Conv1$ in the case of AlexNet. Such evaluation makes more sense than evaluating the summation of all layers because the feature maps from other un-connected layers is not required thus can be discarded during inference time. Comparing with \lq\lq Lin~\etal~\cite{LinFixPoint}\rq\rq, our IFQ-AlexNet networks output 4$\times$ smaller feature maps for $Conv1$ layer  ($55\times 55$ vs. $112\times 112$). Furthermore, our IFQ-AlexNet employs less bits to represent each element of the feature maps of $Conv1$ layer($k=2\, or \,3\, or\, 4$ vs. $\gamma=9\, or\, 10$). Consequently, when comparing IFQ-AlexNet ($k=2$) with \lq\lq Lin~\etal~\cite{LinFixPoint}\rq\rq, our method gives 18$\times$ savings on inference memory for feature maps.

Furthermore, we follow the reference paper~\cite{LinFixPoint} and use Top-5 accuracy to evaluate the performance of the AlexNet-based fixed-point networks. Comparing with \lq\lq Lin~\etal~\cite{LinFixPoint} ($\gamma=9$)\rq\rq, IFQ-AlexNet ($k=2$) gives 2\% improvement accuracy with significant savings on model size and feature maps memory as well. Moreover, comparing the \lq\lq Lin~\etal~\cite{LinFixPoint} ($\gamma=10$)\rq\rq\, and IFQ-AlexNet ($k=4$ ) networks which have higher precision, our method also achieves 2.18$\times$ and 10.9$\times$ savings on model size and feature maps respectively without performance drop.


\subsection{IFQ-Tinier-YOLO face detector}

Face detection has various applications in real life and thus emerges many algorithms such as Faster R-CNN~\cite{FasterRCNN}, SSD~\cite{SSD}, Mask R-CNN~\cite{MaskRCNN} and YOLOv2~\cite{YOLOv2}. In this section, we aim to apply our IFQ-Net to face detection task. For the embedded devices, the simple architecture of a deployed network would give great benefit on the hardware design. Consequently, we make use of YOLOv2 detection algorithm as the framework for our face detector.


We initially employ the Tiny-YOLO~\cite{YOLOv2} network due to its compact size. Furthermore, we design a more compact network Tinier-YOLO based on Tiny-YOLO by: 1) only using half the number of filters in each convolution layer; 2) replacing the $3\times3$ filter into $1\times1$ for the third to last convolution layer; 3)binarizing the weights of all convolution layers by HWGQ. The above three strategies give 4$\times$, ~2$\times$ and 32$\times$ savings respectively and overall 256$\times$ savings on model size resulting in a 246k Bytes face detector.
\begin{table} [!h]
	\centering
	\caption{Comparison on the model size (MB) of the trained face detectors and their detection rate on FDDB dataset~\cite{FDDB}.}\label{tab:facedetector}
	
	\setlength{\tabcolsep}{1.5pt}
	\begin{tabular}{p{1.5cm}<{\centering}|>{\centering}p{1.2cm}|p{1.8cm}<{\centering}|p{1.2cm}<{\centering}|p{1.9cm}<{\centering}}
		
		\hline                           &Tiny-YOLO &{IFQ-Tiny-YOLO$\,$($k$=2)} &Tinier-YOLO &{IFQ-Tinier-YOLO$\,$($k$=2)} \\
		\hline\hline  {model size$\,$(MB)}       & 63.00       & 1.97       &7.89       &\textbf{0.25} \\
		\hline       {detection rate}    &\textbf{0.92}      &0.89             &0.90        &0.84 \\
		\hline
		
	\end{tabular}
\end{table}

\begin{figure}[!ht]
	\begin{center}
		\includegraphics[width=0.85\linewidth]{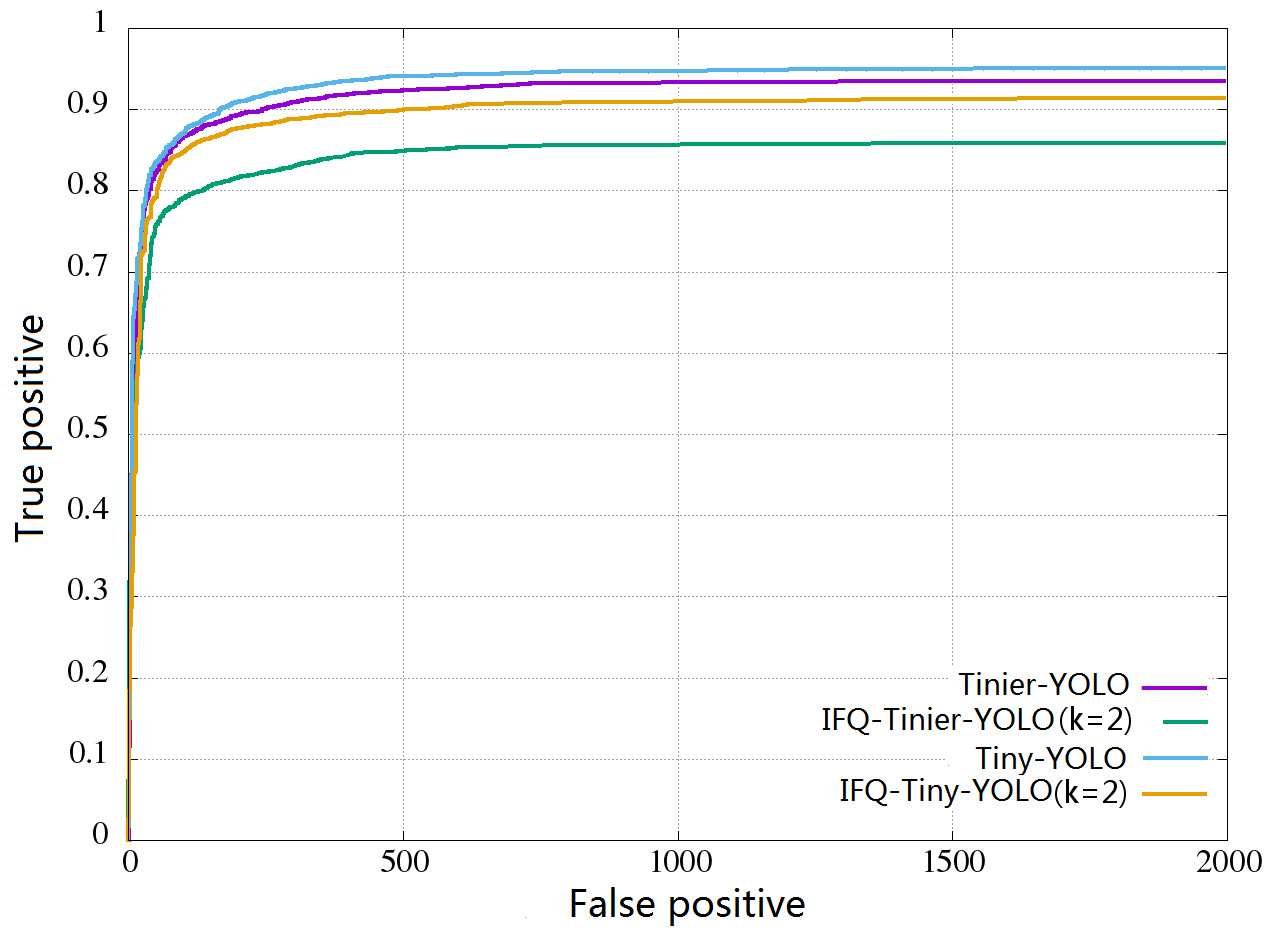}
	\end{center}
	\caption{Performance of the face detectors on FDDB dataset~\cite{FDDB}.}
	\label{fig:compdetector}
\end{figure}

\begin{figure*}[!htb]
	\begin{center}
		\includegraphics[width=0.9\linewidth]{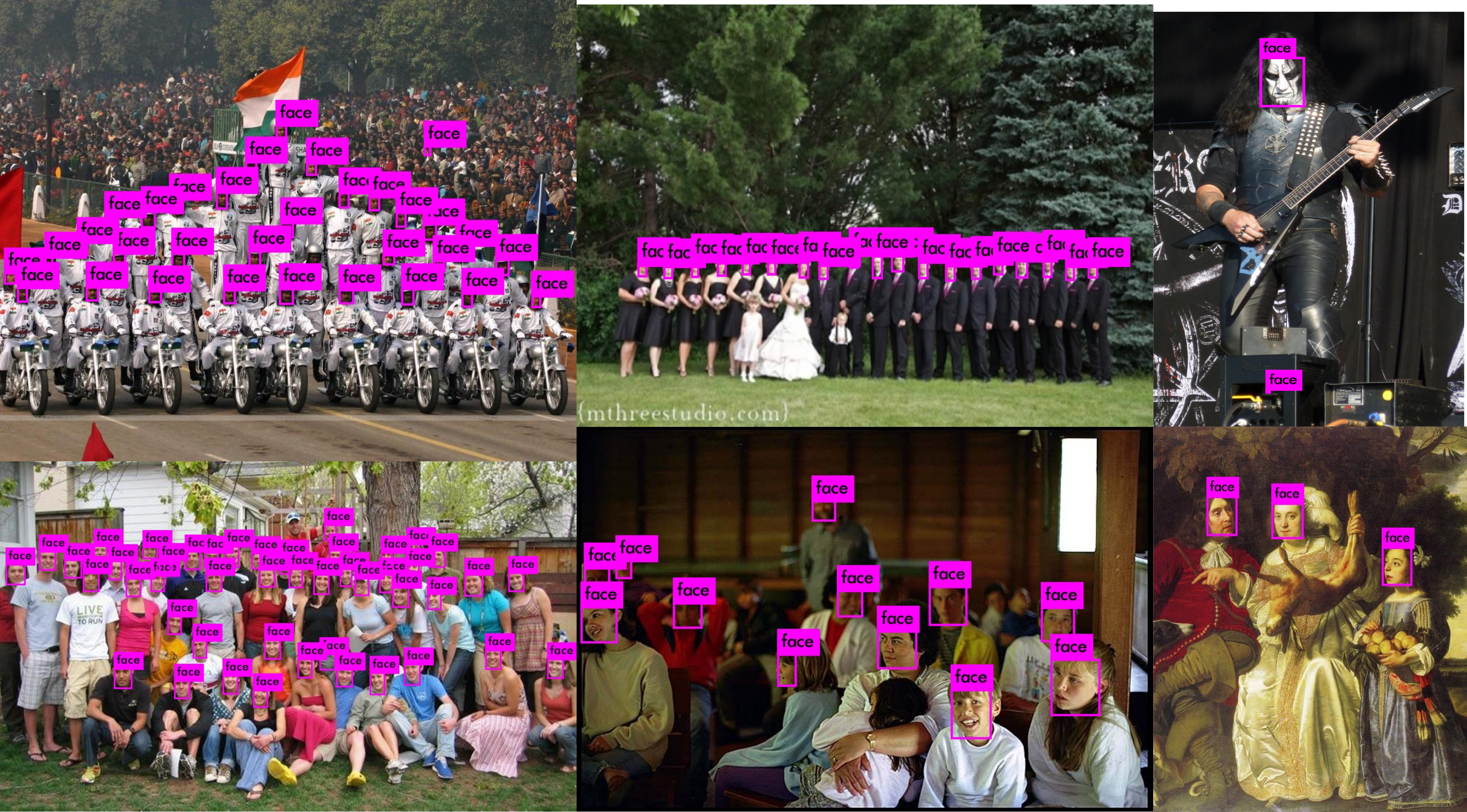}
	\end{center}
	\caption{Qualitative performance of the proposed IFQ-Tiner-YOLO ($k=2$) face detector on Wider Face dataset~\cite{widerface}.}
	\label{fig:visualcheck}
\end{figure*}

We use the training set of Wider Face~\cite{widerface} and Darknet deep learning framework~\cite{darknet} to train the baseline Tiny-YOLO and our Tinier-YOLO networks. Furthermore, to obtain their quantized fixed-point counterparts IFQ-Tiny-YOLO and IFQ-Tinier-YOLO, we first train the quantization network with our HWGQ implementation on Darknet (k = $2$) and then convert each of their substructure into fixed-point. Each network is trained for 100k iterations with batch size 128. The learning rate is initially set to 0.01 and down scaled by 0.1 at $30k$th, $60k$th and $90k$th iteration. Besides, we also inherit the multi-scale training strategy from YOLOv2.

We compare the trained face detectors on FDDB dataset~\cite{FDDB} which contains 5,171 faces in 2,845 testing images. To evaluate the performance of the face detector, we employ detection rate when false positive rate is 0.1 (1 false positive in 10 test images). It corresponds to the true positive rates (y-axis) when the false positive (x-axis) equals to $\lfloor0.1\times2,845\rfloor$ = 284 in Figure~\ref{fig:compdetector}. Such evaluation is more meaningful in real applications when low false positive rate is desired. As illustrated in Table~\ref{tab:facedetector}, comparing with Tiny-YOLO, IFQ-Tiny-YOLO achieves 32$\times$ savings on model size with 3\% drop on detection rate (0.89 vs. 0.92). Furthermore, the proposed IFQ-Tinier-YOLO face detector gives a further 8$\times$ savings over IFQ-Tiny-YOLO with 5\% performance drop. We think its performance is promising in the sense of its extremely compact model size and quite satisfactory detection rate. More importantly, the proposed IFQ-Tinier-YOLO face detector is a fixed-point network which can be easily implemented on embedded devices. The ROC curves of the compared face detectors are illustrated in Figure~\ref{fig:compdetector}. 

Moreover, the proposed IFQ-Tinier-YOLO is also effective on detecting small faces. We test it on Wider Face validation images and show its qualitative results. As shown in Figure~\ref{fig:visualcheck}, our IFQ-Tinier-YOLO also gives nice detection on small faces in various challenging scenarios such as make-up, out of focus, low-illumination, paintings etc.

\section{Conclusions}
In this paper, we presented a novel fixed-point network, IFQ-Net, for embedded vision. It divides a quantization network into substructures and then converts each substructure into fixed-point in either separated or the proposed integrated manner. Especially for the quantized substructures,  which commonly appear in quantization networks, the integrated conversion method removes on-device batch normalization computation, requires no scaling-up effect ($m=0$) yet most importantly does not cause performance drop. We compared our IFQ-Net with the state-of-the-art fixed-point network indicating that our method gives much more savings on model size and feature map memory with similar (or higher in some case) accuracy on ImageNet.

Furthermore, we also designed a fixed-point face detector IFQ-Tinier-YOLO. Comparing with the Tiny-YOLO detector, our model shows its great benefits on embedded devices in the sense of extremely compact model size (246k Bytes), purely fixed-point data operations and quite satisfactory detection rate.

{\small
\bibliographystyle{ieee}
\bibliography{paper_HG}
}

\end{document}